\documentclass{article}
\usepackage{ijcai18}

\usepackage{times}
\usepackage{url}
\usepackage{xcolor}
\usepackage{soul}
\usepackage[utf8]{inputenc}
\usepackage[small]{caption}
\usepackage{epsfig}
\usepackage{graphicx}
\usepackage{amsmath}
\usepackage{amssymb}
\usepackage{helvet}
\usepackage{courier}
\usepackage{dsfont}
\usepackage[ruled,vlined]{algorithm2e}
\usepackage{sansmath}
\usepackage{CJK}
\usepackage{caption}
\usepackage{color}
\usepackage{subfigure}
\usepackage{booktabs}
\usepackage{multirow}
\usepackage[english]{babel}
\usepackage{tabularx}

\newcommand\blfootnote[1]{%
  \begingroup
  \renewcommand\thefootnote{}\footnote{#1}%
  \addtocounter{footnote}{-1}%
  \endgroup
}

\title{Learning to Write Stylized Chinese Characters \\ by Reading a Handful of Examples }

\author{
}

\author{
Danyang Sun$^\ast$, 
Tongzheng Ren$^\ast$, 
Chongxuan Li, 
Hang Su$^\dag$,
Jun Zhu$^\dag$
\\ 
Department of Computer Science and Technology, Tsinghua Lab of Brain and Intelligence\\
State Key Lab for Intell. Tech \& Sys., BNRist Lab, 
Tsinghua University, 100084, China\\
\{sundy16, rtz14, lcx14\}@mails.tsinghua.edu.cn; \{suhangss, dcszj\}@tsinghua.edu.cn
}
\begin{document}

\maketitle

\blfootnote{* equal contribution; $^\dag$ corresponding authors.}

\begin{abstract}
Automatically writing stylized characters is an attractive yet challenging task, especially for Chinese characters with complex shapes and structures. Most current methods are restricted to generate stylized characters already present in the training set, but require to retrain the model when generating characters of new styles. In this paper, we develop a novel framework of Style-Aware Variational Auto-Encoder (SA-VAE), which disentangles the content-relevant and style-relevant components of a Chinese character feature with a novel intercross pair-wise optimization method. In this case, our method can generate Chinese characters flexibly by reading a few examples. 
Experiments demonstrate that our method has a powerful one-shot/few-shot generalization ability by inferring the style representation, which is the first attempt to learn to write new-style Chinese characters by observing only one or a few examples. 
\end{abstract}

\section{Introduction}

The field of character generation remains relatively under-explored compared with the automatic recognition of characters~\cite{pal2004indian,zhong2016handwritten}. This unbalanced progress is disadvantageous for the development of information processing of all languages, especially Chinese, one of the most widely used languages that has its characters incorporated into many other Asian languages, such as Japanese and Korean. 
\begin{figure}[!thb]
    \centering
    \includegraphics[width=1.1\linewidth]{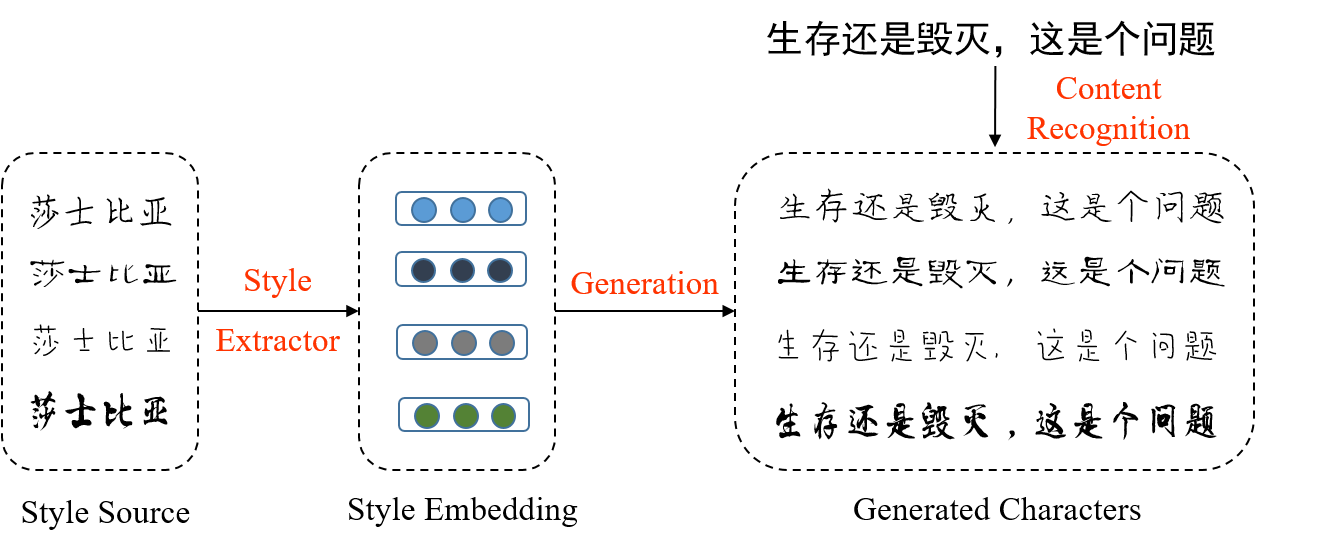}
    \caption{Illustration of Chinese character generation task based on our method. Given a few examples with specified styles (e.g., the signatures), we infer the latent vectors corresponding to different styles. Afterwards, we generate the stylized Chinese characters (``To be or not to be, that is a question'') by recognizing their contents and transferring the corresponding styles (``Shakespeare'') to the targeted characters.}
    \label{example}
\end{figure}
Learning to write stylized Chinese characters by reading a handful of examples (as shown in Fig.~\ref{example}) is also a good testbed for machine intelligence as the task can be finished effectively by humans while still remaining challenging for machines.
Though great efforts have been made on synthesizing simple English or Latin characters~\cite{graves2013generating,lake2015human}, only a little exploration has been done on Chinese character generation. This is not surprising as Chinese characters pose new challenges due to their complexity and diversity. First, Chinese characters have a much larger dictionary compared with the alphabetic writings. For instance, there are 27,533 unique Chinese characters in the official GB18030 charset, with daily used ones up to 3,000. Recently, \citeauthor{zhang2017drawing}~[2017] propose to generate Chinese characters based on Recurrent Neural Networks~\cite{hochreiter1997long}, which learns a low dimensional embedding as the character label to solve the issue of large dictionary. 
However, this method only generates characters without any style information, and heavily relies on the online temporal information of the strokes, which cannot generalize to the off-line applications.

Moreover, Chinese characters have more complex shapes and structures than other symbolic characters such as Latin alphabets. 
Some attempts have been made on Chinese character generation by assembling components of radicals and strokes~\cite{xu2009automatic,zong2014strokebank,lian2016automatic}. However, these models rely on the preceding parsing, which requires each character to have a clear structure. Therefore, they cannot deal with a joined-up writing style , in which strokes are connected and cursive.

Finally, the generation of stylized Chinese characters is also challenged by the complexity in writing styles. The recent ``Rewrite''~\cite{kaonash2016rewrite} and its advanced version  ``zi2zi''~\cite{kaonash2017zi2zi} based on the ``pix2pix''~\cite{Isola2016ImagetoImageTW} implement font style transfer of Chinese characters by learning to map the source style to a target style with thousands of character pairs. However, these methods have to re-train a model from scratch when observing some Chinese characters with a new style, failing to re-utilize the knowledge learned before. 

\subsection{Our Proposal}

To address the aforementioned challenges, we propose a novel framework named Style-Aware Variational Auto-Encoder (SA-VAE) to conduct the stylized Chinese character generation. Compared with the alternative popular deep generative model of Generative Adversarial Networks (GAN)~\cite{goodfellow2014generative}, Variational Auto-Encoder (VAE)~\cite{kingma2013auto} naturally has the ability of posterior inference to reveal the underlying factors, usually yielding better inference effects than inference extensions to GAN~\cite{donahue2016adversarial,dumoulin2016adversarially}. In this paper, we make a feature-disentangled extension to VAE both in the model level and algorithm level.

It is imperative to extract a pure and accurate style representation for the purpose of learning a new writing style given one or a few examples, which therefore drives us to disentangle the style and content features from the given samples. To our best knowledge, the most relevant study to solve this two-factor disentanglement problem is using a bilinear model~\cite{tenenbaum1997separating}. However, the disentanglement based on matrix decomposition limits its applicabilities in the complicated situations with highly nonlinear combinations. 
Therefore, in the model level, we leverage the powerful capabilities of deep learning for non-linear function approximation to design a style inference network and a content recognition network, which encode the style and content information respectively. Meanwhile, in the algorithm level, we propose a novel intercross pair-wise optimization method, which encourages the extracted style embedding more pure and reliable.

Unlike the general image generation like faces~\cite{liu2015faceattributes} or bedrooms~\cite{yu2015lsun}, Chinese character generation is also challenged by the large dictionary and rigorous structures. It may lead to confusing or incorrect results even if a small bias is introduced in Chinese character generation. To address this issue, we integrate the character knowledge into our framework by considering the specific configuration and radicals information of Chinese characters, which offers significant benefits when we generate characters with complex structures. In order to generate characters with an unseen style, our model learns the characteristics of numerous printing and handwritten styles as our style bank. Relying on the powerful generalization capabilities of our model, the proposed model makes a reasonable inference and conducts effective generalization on new styles without the need to retrain the model.    

\begin{figure*}[!thb]
    \centering
    \includegraphics[width=1\linewidth]{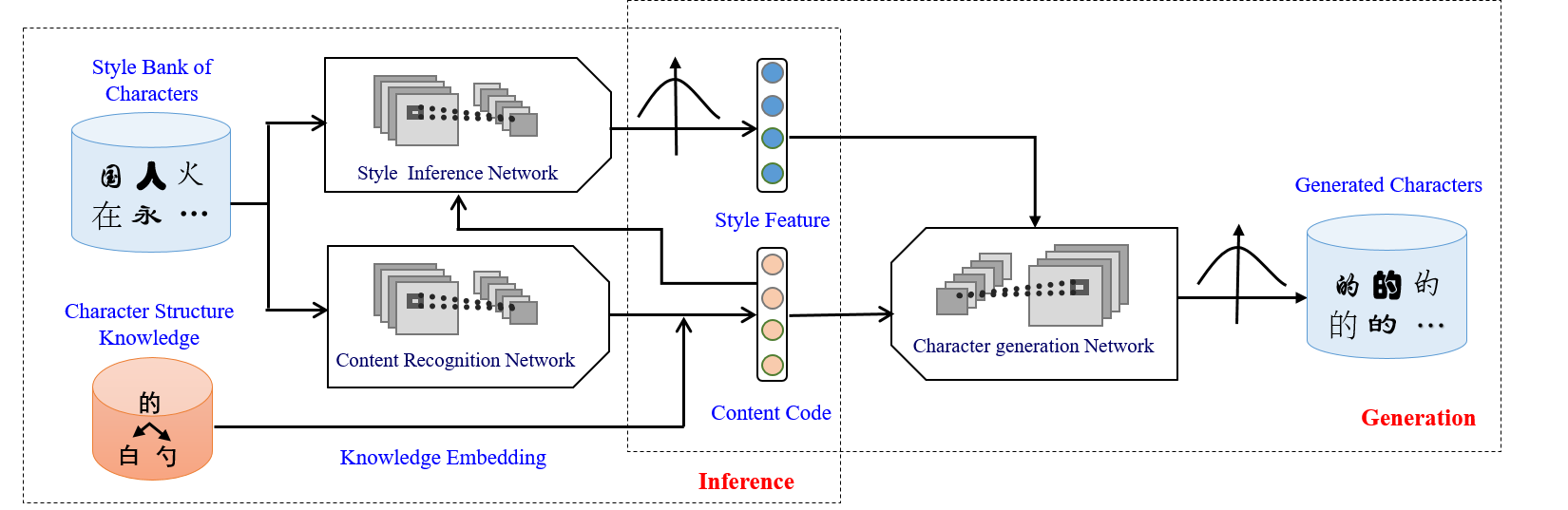}
    \caption{Our proposed SA-VAE framework mainly consist of three subnets, including a Content Recognition Network $\mathcal{C}$, a Style Inference Network $\mathcal{S}$ and a Character Generation Network $\mathcal{G}$. $\mathcal{S}$ and $\mathcal{C}$ extract the style feature and the content feature respectively and $\mathcal{G}$ combines these two features to conduct the generation. Besides, we introduce the domain knowledge of Chinese characters $\mathcal{K}$ to get a more informative content representation. The training process is executed by an intercross pair-wise way.}
    \label{model}
\end{figure*}

Extensive experiments demonstrate that our method can generate stylized Chinese characters by reading only one or a few samples with diverse styles. To the best of our knowledge, this is the first attempt to learn to write Chinese characters of new styles in this one-shot/few-shot scenario. To sum up, our major contributions are as follows:
\begin{itemize}
\item We propose a novel intercross pair-wise optimization method to disentangle the style and content features of Chinese characters, which is a general technique to solve two-factor disentanglement problems with supervision.
\item We introduce the domain knowledge of Chinese characters into our model \emph{a priori} to guide the generation. 
\item Our proposed framework (SA-VAE) can achieve desirable style inference and generate stylized Chinese characters in the one-shot/few-shot setting, even for the unseen styles in the training stage.
\end{itemize}

\section{Methodology}
\label{sec:2}
In this section, we first formally present our task and the architecture of the whole framework. Then, we present our intercross pair-wise optimization method in detail. Finally, we show how to conduct one-shot/few-shot stylized Chinese character generation with our model.

\subsection{Problem Formulation}

To frame the task formally, we make the mild assumption that all Chinese characters are independently decided by the content factor and the style factor, which can be denoted as:
\begin{equation}
\label{eq:assumption}
\mathbf{x}_{i,j}\leftarrow (\mathbf{s}_i, \mathbf{c}_j) ,
\end{equation}
where \(\mathbf{s}_i\) represents style \(i\), \(\mathbf{c}_j\) represents content \(j\), and \(\mathbf{x}_{i,j}\) is the character with style \(i\) and content \(j\).  

Based on this assumption, we propose to infer the style representation when observing the characters, and integrate these style features with the content features to generate target characters. As the characters do not have an explicit separation between style and content, we thus have to do ``reverse-engineering"---
recognize the content and disentangle the style information from given characters, and combine the style with target contents to generate the stylized Chinese characters.

To this end, we first build a dataset containing all the Chinese characters with a diverse set of styles as our style-bank, which is denoted as:
\begin{equation}
\mathcal{X} =\{\mathbf{x}_{i,j}\} , i=1,2,...,M , j=1,2,...,N  ,
\end{equation}
where $M$ is the number of styles and $N$ is the total number of unique Chinese characters in the training dataset (See Sec.~\ref{subsec:data_preparation} for details). 

Afterwards, we build a Content Recognition Network $\mathcal{C}$ and a Style Inference Network $\mathcal{S}$ to respectively obtain the disentangled features of content \(\mathbf{c}_j\) and style \(\mathbf{s}_i\) for each Chinese character \(\mathbf{x}_{i,j}\). The characters are then generated via a generation network $\mathcal{G}$. In the one-shot/few-shot setting, we allow the model to observe only one or a few characters with a new style denoted as $\{\mathbf{x}_{M+k , l:l+m}\} $ where $k > 0, m \geq0$.
Based on these characters, the style feature \(\mathbf{s}_{M+k}\) can be inferred by $\mathcal{S}$, which is combined with any content feature \(\mathbf{c}_n \in \{\mathbf{c}_{1:N}\}\) to generate the desirable stylized Chinese character \(\mathbf{x}_{M+k, n}\) via \(\mathcal{G}\).


\subsection{Model Architecture} 

Here, we present the details of our model. The model mainly consists of three modules of a Content Recognition Network $\mathcal{C}$, a Style Inference Network $\mathcal{S}$ and a Character Generation Network $\mathcal{G}$ as illustrated in Fig.~\ref{model}. The whole process can be separated as two stages --- {\it inference} and {\it generation}. In the inference stage, we first learn to disentangle the latent features into content-related and style-related components based on the Content Recognition Network and Style Inference Network respectively.
In the generation stage, we take the content and style vectors as inputs via a deconvolutional network, such that the stylized characters are well reconstructed with the style feature that is appropriately captured in the inference stage. The training process is executed by an intercross pair-wise way (See Sec.~\ref{subsec:intercross}) for a reliable 
disentanglement.\\

\noindent\textbf{Content Recognition Network}. Content recognition aims to recognize the content of every Chinese character image, which provides a correct content label to the style inference network and generation network. We refer to the previous works on state-of-the-art character recognition ~\cite{xiao2017building,zhong2016handwritten} to satisfy our need and formulate it as:  
\begin{equation}
\mathcal{C}: \mathbf{y} = f_{\eta}(\mathbf{x}) ,
\end{equation}
where $\mathbf{x}$ denotes the character image, $\mathbf{y}$ is the content label, and $\eta$ denotes the network's parameters.\\


\noindent\textbf{Character Structure Knowledge}. Utilizing the domain knowledge of Chinese characters is essential to develop a compact model and improve the generation quality. 
 Previous deep generative models~\cite{radford2015unsupervised,li2017max} usually cast each category as a one-hot embedding, and concatenate it to all the feature maps in every layer. This method works well for some small datasets such as MNIST and SVHN which have only ten classes but fails to solve the Chinese character generation with thousands of classes\footnote{The number of classes is comparable to that of the ImageNet.}. It may lead to an explosive growth of model parameters. To address this issue, we propose a more informative character encoding method by introducing the configuration and radical information of Chinese characters into our framework and build a corresponding index table $\mathcal{K}$. 

Compared with the one-hot embedding, our encoding method can reuse the configuration and radical information shared among all the Chinese characters. The one-hot embedding $y$ of every character can be transformed to a unique content hashing code through a index table $\mathcal{K}$ as
\begin{equation}
\mathcal{K} : \mathbf{c} = T[\mathbf{y}] .
\end{equation}

\begin{figure}
    \centering
    \includegraphics[width=1\linewidth]{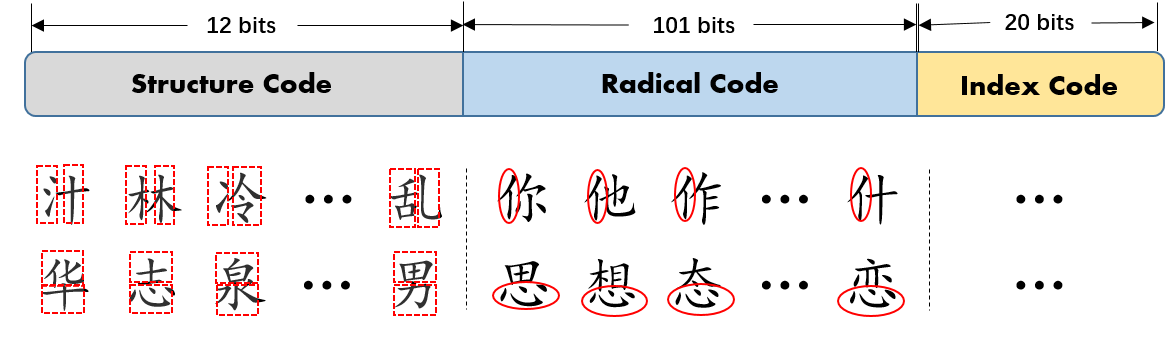}
    \setlength{\abovecaptionskip}{-1em}
    \caption{A diagram to illustrate our encoding method, which contains the configuration information and the radical information.}
    \label{fig:code}
\end{figure}

Our encoding method is shown in Fig.~\ref{fig:code}, which is much shorter than the one-hot embedding (3000 bits in total). The first 12 bits are used to identify the 12 common configurations in Chinese characters, including {\it up-down}, {\it left-right}, {\it surroundings} and so on. The middle 101 bits are used to identify 100 frequently used radicals and the case of missing radicals.
The last 20 bits are as a binary index when the former two parts are the same. Proved by the experimental results, our proposed encoding method compresses the embedding dimension while does not harm the model's capacity compared with other embedding methods.\\

\noindent \textbf{Style Inference Network}. The Style Inference Network is the key module to make our framework have the capacity of achieving accurate style inference. To automatically disentangle the style feature from the content information, we feed the content hashing code $\mathbf{c}$ to this network besides the character image $\mathbf{x}$. 
It is realized as a diagonal Gaussian distribution parameterized by the convolutional neural network. Hence, we formulate the Style Inference Network as: 
\begin{equation}
\mathcal{S}: q_{\phi}(\mathbf{s} | \mathbf{x} , \mathbf{c}) = \mathcal{N}(\mathbf{s} | \mathbf{\mu}_{\phi(\mathbf{x} , \mathbf{c})} , \text{diag}(\mathbf{\sigma}^2_{\phi(\mathbf{x} , \mathbf{c})})) ,
\end{equation}
where $\mathbf{\mu}_{\phi(\mathbf{x} , \mathbf{c})}$ and $\mathbf{\sigma}^2_{\phi(\mathbf{x} , \mathbf{c})} $ correspondingly denote the mean and variance of the Gaussian distribution, and $\phi$ denotes the parameters of the convolutional neural network. 

On the other hand, for a reliable disentanglement, a good Style Inference Network should try its best to 
satisfy this equality:
\begin{equation}
\label{eq:target}
\begin{aligned}
q_\phi(\mathbf{s}_i|\mathbf{x}_{i,j},\mathbf{c}_j)=q_\phi(\mathbf{s}_i| \mathbf{x}_{i,k},\mathbf{c}_k),
\end{aligned}
\end{equation}
which requires the Style Inference Network to provide the same posterior distribution given different characters with the same style. This target implicitly ensures that the Style Inference Network can extract the style information because the content information of the inputs are different. This property is very important as our main purpose is to disentangle the style information and achieve one-shot inference. We will explain how to achieve Eq.~\eqref{eq:target} by the intercross pair-wise optimization method in detail in Sec.~\ref{subsec:intercross}.\\

\noindent \textbf{Character Generation Network}. In order to generate different characters with diverse styles flexibly, the Character Generation Network takes the content hashing code $\mathbf{c}$ and the style feature $\mathbf{s}$ as two independent inputs, and these two disentangled factors jointly decide the generated characters.
We use a Bernoulli distribution parameterized with a deconvolutional neural network~\cite{zeiler2010deconvolutional,radford2015unsupervised} to model the binary image of Chinese characters, which can be formulated as:
\begin{equation}
\mathcal{G}: p_\theta(\mathbf{x} | \mathbf{s} , \mathbf{c}) = \text{Bern}(\mathbf{x} | \mathbf{\mu}_{\theta(\mathbf{s} , \mathbf{c})}) ,
\end{equation}
where  $\mathbf{\mu}_{\theta(\mathbf{s} , \mathbf{c})}$ denotes the logits of Bernoulli distribution and $\theta$ denotes the parameters of deconvolutional neural network.\\


\subsection{Intercross Pairwise Optimization}
\label{subsec:intercross}
Generally, latent variable generative models are usually trained by maximizing the marginal log-likelihood, but it is  often intractable to implement the optimization directly. To address this issue, variational inference~\cite{wainwright2008graphical} can be applied by optimizing a variational lower bound parameterized with \(q_\phi(\mathbf{z}|\mathbf{x})\) instead of the log-likelihood.




In our Chinese character generation task, the style and content of Chinese characters are independent as in Eq.(\ref{eq:assumption}). Meanwhile, the content code $\mathbf{c}$ can be obtained by the well-trained $\mathcal{C}$. Therefore, we can get the conditional variational lower bound of vanilla VAE:
\begin{align}\label{condition lower bound}
\log p_\theta(\mathbf{x}|\mathbf{c}) \geq & \log p_\theta(\mathbf{x}|\mathbf{c}) -\mathds{KL}\left[q_{\phi}(\mathbf{s|x,c})\| p_\theta(\mathbf{s|x,c})\right] \nonumber \\
  =&\mathcal{L}_{ELBO}(\mathbf{x};\mathbf{c},\theta,\phi)  ,
  \end{align}
where $\mathds{KL}$ denotes the Kullback Leibler Divergence and $p_\theta(\mathbf{s|x,c})$ is the true posterior from the generative model.

\begin{figure}
    \centering
    \includegraphics[width=1\linewidth]{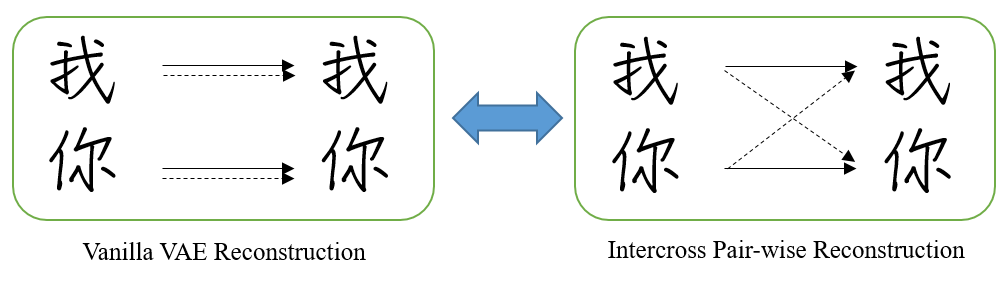}
    \caption{The comparison between vanilla VAE and our intercross pair-wise training, in which dotted lines and solid lines denote the style providers and the content providers, respectively.}
     \label{fig:training}
\end{figure}

However, directly optimizing the Eq.~\eqref{condition lower bound} cannot ensure that characters with the same font share a similar style embedding. This is because the vanilla variational inference treats each character separately and there is no strong correlation between different characters, especially those with the same style. 
Hence, we propose a new objective of intercross pairwise optimization, which is still a lower bound of the log-likelihood. Instead of reconstructing the input as in the vanilla VAE, we try to produce the target character based on two characters with the same style information but different contents, that is to say, the generated character obtains the style information by another character with the same style, as is illustrated in Fig.~\ref{fig:training}.

Specifically, we consider a pair of characters $(\mathbf{x}, \mathbf{x}^{\prime})$, which share the same style $\mathbf{s}$ but with different contents $\mathbf{c}$ and $\mathbf{c}^{\prime}$. We define the objective function as   
\begin{align}
\tilde{\mathcal{L}}(\mathbf{x}, \mathbf{x}^{\prime}, &\mathbf{c}, \mathbf{c}^{\prime};\theta , \phi) \nonumber \\ 
=& \log p_\theta(\mathbf{x}|\mathbf{c})-\mathds{KL}[q_{\phi}(\mathbf{s}|\mathbf{x^\prime}, \mathbf{c^\prime})\|p_\theta(\mathbf{s}|\mathbf{x},\mathbf{c})]. \label{eq:analyze} 
\end{align}

By maximizing this objective, we can achieve the target of Eq.(\ref{eq:target}) which is the key point of disentanglement and one-shot inference. Specifically, from Eq.(\ref{eq:analyze}), we know that maximizing $\tilde{\mathcal{L}}$ is equivalent to minimizing the $\mathds{KL}$ term, since the data likelihood is invariant.  Meanwhile, as we randomly sample the training pairs, the dual pairs will also appear. Hence, it simultaneously minimizes:
\begin{align}
\label{eq:a}
\mathds{KL}[q_{\phi}(\mathbf{s}|\mathbf{x^\prime}, \mathbf{c^\prime})\|p_\theta(\mathbf{s}|\mathbf{x}, \mathbf{c})], \\ 
\mathds{KL}[q_{\phi}(\mathbf{s}|\mathbf{x},\mathbf{c})\|p_\theta(\mathbf{s}|\mathbf{x^\prime}, \mathbf{c^\prime})].
\end{align}
With the assumption in Eq.(\ref{eq:assumption}), we know that the real posterior distribution of the style is identical for $\mathbf{x}$ and $\mathbf{x^\prime}$, that is
\begin{equation}
\label{eq:condition}
p_\theta(\mathbf{s}|\mathbf{x},\mathbf{c})=p_\theta(\mathbf{s}| \mathbf{x^\prime},\mathbf{c^\prime}) .
\end{equation}
\noindent Ideally, if two $\mathds{KL}$ terms in Eq.(10) and Eq.(11) are optimized to zero, the point of equilibrium is arrived and Eq.(\ref{eq:target}) holds by substituting Eq.(\ref{eq:condition}). Therefore, this derivation explains why our objective function can achieve a reliable disentanglement.


Furthermore, this objective function can be rewritten as:
\begin{align}\label{eq:reconstruction}
\tilde{\mathcal{L}}(\mathbf{x}, &\mathbf{x}^{\prime}, \mathbf{c}, \mathbf{c}^{\prime};\theta , \phi) \\ 
=&\mathds{E}_{q_{\phi}(\mathbf{s}|\mathbf{x^{\prime}},\mathbf{c^{\prime}})}[\log p_{\theta}(\mathbf{x}|\mathbf{s},\mathbf{c})]-\mathds{KL}[q_{\phi}(\mathbf{s}|\mathbf{x^\prime}, \mathbf{c^\prime})\|p_\theta(\mathbf{s})],  \nonumber
\end{align}

\begin{algorithm}[!htb]
	\DontPrintSemicolon
	\caption{Training Algorithm}
	\KwIn{Chinese characters dataset $\mathbf{X} =\{\mathbf{x}_{i,j}\}$}
	\KwOut{The model parameters : $\mathbf{\theta}$ , $\mathbf{\phi}$ , $\mathbf{\eta}$}
	Randomly initialize  $\mathbf{\theta}$ , $\mathbf{\phi}$ , $\mathbf{\eta}$.\\
    $\mathbf{\eta} \leftarrow $ Pre-train a Content Recognition Network $\mathcal{C}$ 
    
	\Repeat{Converge}
	{
		 $\mathbf{x} \leftarrow $ Randomly select a mini-batch from $\mathbf{X}$ \\
         $\mathbf{x}^{\prime} \leftarrow $ Randomly select a mini-batch correspondingly shares the same font with $\mathbf{x}$ \\
         $\mathbf{c} \leftarrow $ Get the content code of $\mathbf{x}$ using $\mathcal{C}$ and $\mathcal{K}$\\
         $\mathbf{c}^{\prime} \leftarrow $ Get the content code of $\mathbf{x}^{\prime}$ using $\mathcal{C}$ and $\mathcal{K}$\\
		 $\mathbf{g} \leftarrow $ $\nabla_{\theta , \phi} \mathcal{L}( \mathbf{x}, \mathbf{x}^{\prime}, \mathbf{c}, \mathbf{c}^{\prime};\theta , \phi)$  (Gradient of (\ref{eq:reconstruction}))\\
         $\mathbf{\theta}, \mathbf{\phi} \leftarrow $ Update parameters using gradients $\mathbf{g}$
	}
    \label{alg:learning}
\end{algorithm}

\noindent where $p_\theta(\mathbf{s})$ denotes the prior distribution of $\mathbf{s}$. This form is more explicit to derive our training algorithm, in which the first term means a reconstruction loss and the second term means a regularization. 
Note that our proposed intercross pair-wise optimization method is a general technique for two-factor disentanglement problems~\cite{tenenbaum1997separating}, which is not only restricted in the task of stylized Chinese character generation. The details of the intercross pair-wise optimization is shown in Alg.~\ref{alg:learning}.


\subsection{One-shot/Few-shot Character Generation}

When testing with one-shot setting, we observe a new style with only one character, then we model other characters using the following way:
\begin{equation}
\begin{aligned}
p( \mathbf{x}_{i,1:n}&|\mathbf{c}_{1:n},  \mathbf{x}_{i,j}) = \\
\int & {p_{\theta}(\mathbf{x}_{i,1:n}|\mathbf{c}_{1:n} , \mathbf{s}_i) q_{\phi}(\mathbf{s}_i|\mathbf{x}_{i,j}, T[f_{\eta}(\mathbf{x}_{i,j})]) d{\mathbf{s}_i}} ,
\end{aligned}
\end{equation}
in which \(\mathbf{x}_{i,j}\) is the observed character as the style-provider, \(\mathbf{c}_j\) is the content code of \(\mathbf{x}_{i,j}\) which can be obtained from Content Recognition Network \(\mathcal{C}\), \(\mathbf{s}_i\) is the new style embedding inferred by our Style Inference Network \(\mathcal{S}\) from the character \(\mathbf{x}_{i,j}\), \(\mathbf{c}_{1:n}\) means the content codes of characters we want to generate and \(\mathbf{x}_{i,1:n}\) are the target characters with the new style generated by Character Generation Network \(\mathcal{G}\).


If we can observe a few more characters, which means testing with few-shot setting, we use the averaged embedding vectors:
\begin{equation}
\begin{aligned}
\hat{\mathbf{s}_i} =\frac{1}{n} \sum_{j=1}^{n}\mathbf{s}_{i}^{(j)},\;
\textrm{where } \mathbf{s}_{i}^{(j)} \sim q_\phi(\mathbf{s}_i|\mathbf{x}_{i,j}, \mathbf{c}_j), 
\end{aligned}
\end{equation}
where $j = 1 ... n$ with \(n\) being the number of observed characters; \(\mathbf{x}_{i,j}\) are corresponding characters we observed and \(\mathbf{c}_j\) are the content code of \(\mathbf{x}_{i,j}\). \(\mathbf{s}_i^{(j)}\) are independent random variables correspondingly obey the posterior distribution \(q_\phi({\mathbf{s_i}|\mathbf{x_{i,j}}, \mathbf{c_j}})\).
With Eq.(\ref{eq:target}), we can get ${Var}[\hat{\mathbf{s}_i}] = \frac{1}{n}{Var}[\mathbf{s}_i]$,
which means that we can obtain more stable estimation of the style information by averaging over multiple characters.



\section{Experiments}
\label{sec:exp}
In this section, we first introduce the dataset we build. Then we present the one-shot/few-shot generation results and  present some experimental results to verify that our model has a strong ability of style generalization and disentanglement. Finally, we show the necessity and advantage of using the proposed encoding method with human knowledge.

\subsection{Data Preparation}
\label{subsec:data_preparation}
As our main purpose is to generalize over new styles based on learning sufficient support styles, we need to get enough styles for our model to learn. As no existing datasets satisfy our goal, we build a new one. The dataset consists of Chinese characters in 200 styles collected from the Internet. We randomly select $80\%$ of these styles (i.e., 160 styles) to be used in the style-bank for our training and the rest $20 \%$ (i.e., 40 styles) to be used for test. We choose the most frequently used 3000 Chinese characters from each style in the style-bank to build our training set. 

\subsection{One-shot/Few-shot Generation Results}
\begin{figure}
\centering
\subfigure[An excerpt from an classical essay Lou Shi Ming.]{
\begin{minipage}[b]{\linewidth}
\centering
\includegraphics[width=0.55\linewidth]{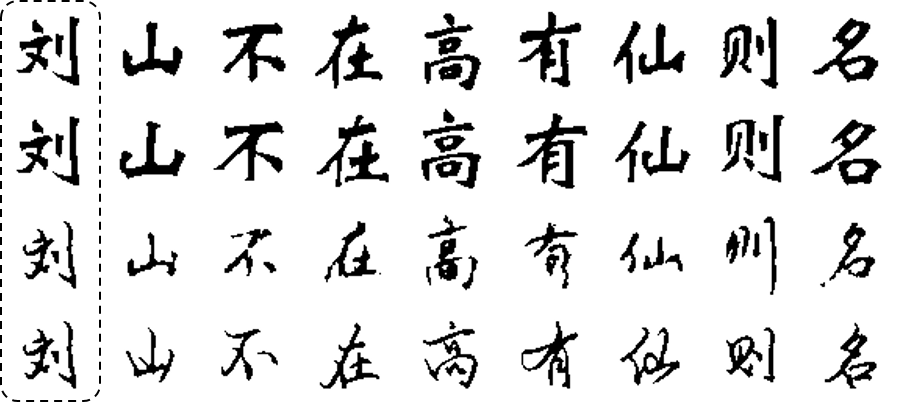}
\end{minipage}
}
\subfigure[An excerpt from the Analects of Confucius.]{
\begin{minipage}[b]{\linewidth}
\centering
\includegraphics[width=0.65\linewidth]{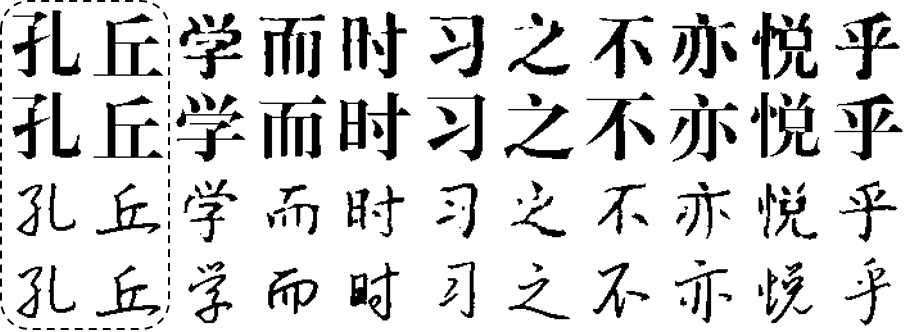}
\end{minipage}
}
 \caption{The generation results in the one-shot and few-shot setting. The characters in the first and third rows are generated by our SA-VAE framework compared with the ground truth in the second and fourth rows. The characters in the dotted frame are used to specified a a new targeted style for the model.} 
 \label{fig:fewshot}
\end{figure}

Fig.~\ref{fig:fewshot} shows the one-shot and few-shot generation results. Both results are presented with a printing style and a handwritten style, which demonstrates that our model has a capacity of dealing with diverse styles.

Meanwhile, we make a comparison with some baseline models of font style transfer ("Rewrite"~\cite{kaonash2016rewrite}, "zi2zi"~\cite{kaonash2017zi2zi}) as shown in Table~\ref{tab:1}. As the former setting, our SA-VAE generates the new-stylized characters in the table by observing only one character. As for other methods, if the target style is in the training set (1000 samples used to train here), the results are close to ours. However, if the target style is new (unseen in the training stage) and the amount of style providers is small (10 characters with the new style used to finetune here), they hardly capture anything about the new style. Therefore, our SA-VAE presents a clear advantage in the one-shot/few-shot inference. 

\begin{table}

\centering
\begin{tabular}{c|c|c}
\toprule
Setting& Method& Results\\
\midrule
\multirow{4}{3cm}[-0.25cm]{\quad \;\; Seen style\\(1000 characters)}
& Rewrite & \begin{minipage}{2.5cm} \centering \includegraphics[width=1\linewidth]{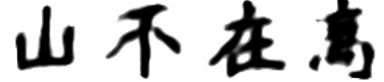} \end{minipage} \\
& zi2zi & \begin{minipage}{2.5cm} \centering \includegraphics[width=1\linewidth]{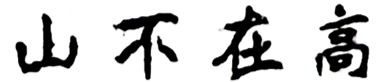} \end{minipage} \\
& \textbf{SA-VAE} & \begin{minipage}{2.5cm} \centering \includegraphics[width=1\linewidth]{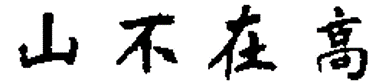} \end{minipage} \\
&Ground Truth & \begin{minipage}{2.5cm} \centering \includegraphics[width=1\linewidth]{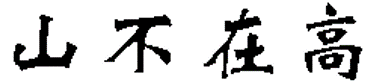} \end{minipage} \\
\midrule
\multirow{4}{3cm}[-0.25cm]{\quad \;\;\; New style\\(Extra 10 characters)}
& Rewrite &  \begin{minipage}{2.5cm}\centering \includegraphics[width=1\linewidth]{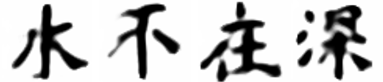}\end{minipage} \\
& zi2zi &  \begin{minipage}{2.5cm}\centering \includegraphics[width=1\linewidth]{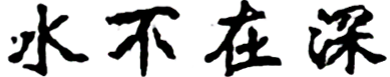}\end{minipage} \\
& \textbf{SA-VAE} &  \begin{minipage}{2.5cm} \vspace{0.1cm} \centering \includegraphics[width=1\linewidth]{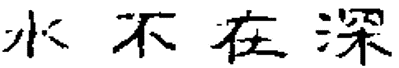}  \end{minipage} \\
&Ground Truth &  \begin{minipage}{2.5cm} \vspace{0.1cm} \centering \includegraphics[width=1\linewidth]{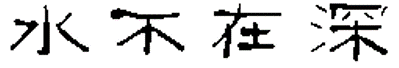}\end{minipage} \\
\bottomrule
\end{tabular}
\captionsetup{justification=centering}
\caption{Comparison of generation results obtained by our SA-VAE and other font style transfer methods in different settings.}
\label{tab:1}
\end{table}





\subsection{Style Generalization}
\begin{figure}
\centering

\subfigure[Interpolation]{
\begin{minipage}[b]{0.55\linewidth}
\centering
\includegraphics[width=\linewidth]{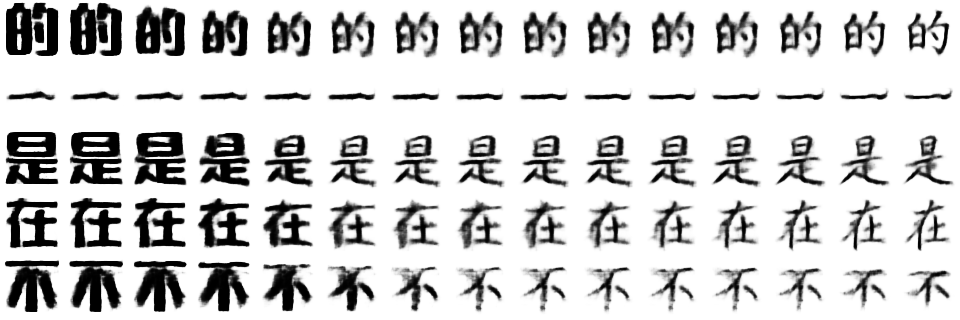}
\end{minipage}
}
\subfigure[NN results]{
\begin{minipage}[b]{0.35\linewidth}
\includegraphics[width=1\linewidth]{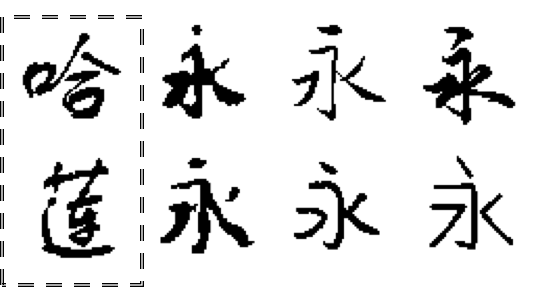}
\end{minipage}
}
 \caption{(a) is the interpolation of two styles. (b) is the nearest neighbor of one-shot generation. The first characters in the dotted frame are style sources and the second ones are generated. The third and the forth are the two nearest neighbors in the training set.} 
 \label{fig:generalization}
\end{figure}

The style generalization is the key point in our SA-VAE framework, which proves our model has a capacity of generalizing to a new style. To show that our model can achieve this goal, we do the interpolation between two disparate styles. Besides, we also find the nearest neighbors of our one-shot generalization results to show our method is not finding the most similar style in the training set. The results are shown in Fig.~\ref{fig:generalization}.

\subsection{Evaluation of Disentanglement}

\begin{figure}
    \centering
    \includegraphics[width=1\linewidth]{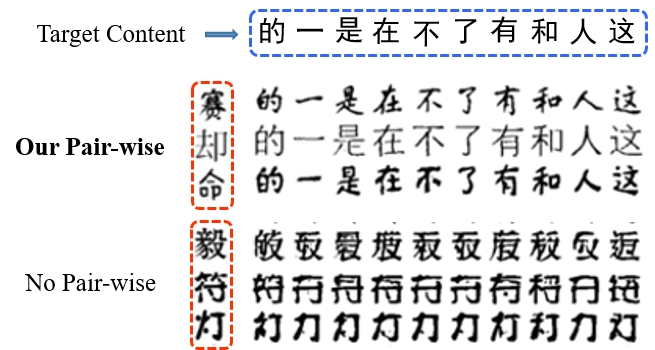}
    \caption{We show the one-shot generation results produced by whether using proposed intercross pair-wise optimization or not.}
    \label{fig:disentangle}
\end{figure}

Successfully disentangling the style information from the content information is another necessity to achieve one-shot generation with given styles. First, we show a qualitative experimental result as shown in Fig~\ref{fig:disentangle}. We can see compared with our SA-VAE, the style embedding extracted without intercross pair-wise optimization method contains too much content information and leads to blurry results.

\begin{table}[!thb]

\centering
\begin{tabular}{c|c|c}
\toprule
Methods & Style Acc & Content Acc \\
\midrule
Vanilla VAE optimization & 43.44\% & 28.55 \%\\
\textbf{Our Pair-wise optimization} & \textbf{47.42}\% & \textbf{1.28} \% \\
\bottomrule
\end{tabular}
\caption{Classification accuracy using the style embedding}
\label{tab:classification}
\end{table}

Meanwhile, We also conduct a quantitative experiment. Based on the intuition that a pure style embedding should contain enough information to represent its style while little information to represent its content, we choose a simple classifier to do some auxiliary classification tasks using the style embeddings obtained by our intercross pair-wise optimization method and vanilla VAE's method.
To simplify the training, we only use 100 unique characters per style as the training data to conduct style classification and use 100 styles per character to conduct character content classification, then test the prediction accuracy over the rest of samples. The results are shown in Table \ref{tab:classification}.


Both the qualitative and quantitative experiment results provide a strong proof that our intercross pair-wise optimization method has a powerful ability of disentanglement.

\subsection{Structure Knowledge of Chinese Characters}
\begin{figure}[!thb]
    \centering
    \includegraphics[width=1\linewidth]{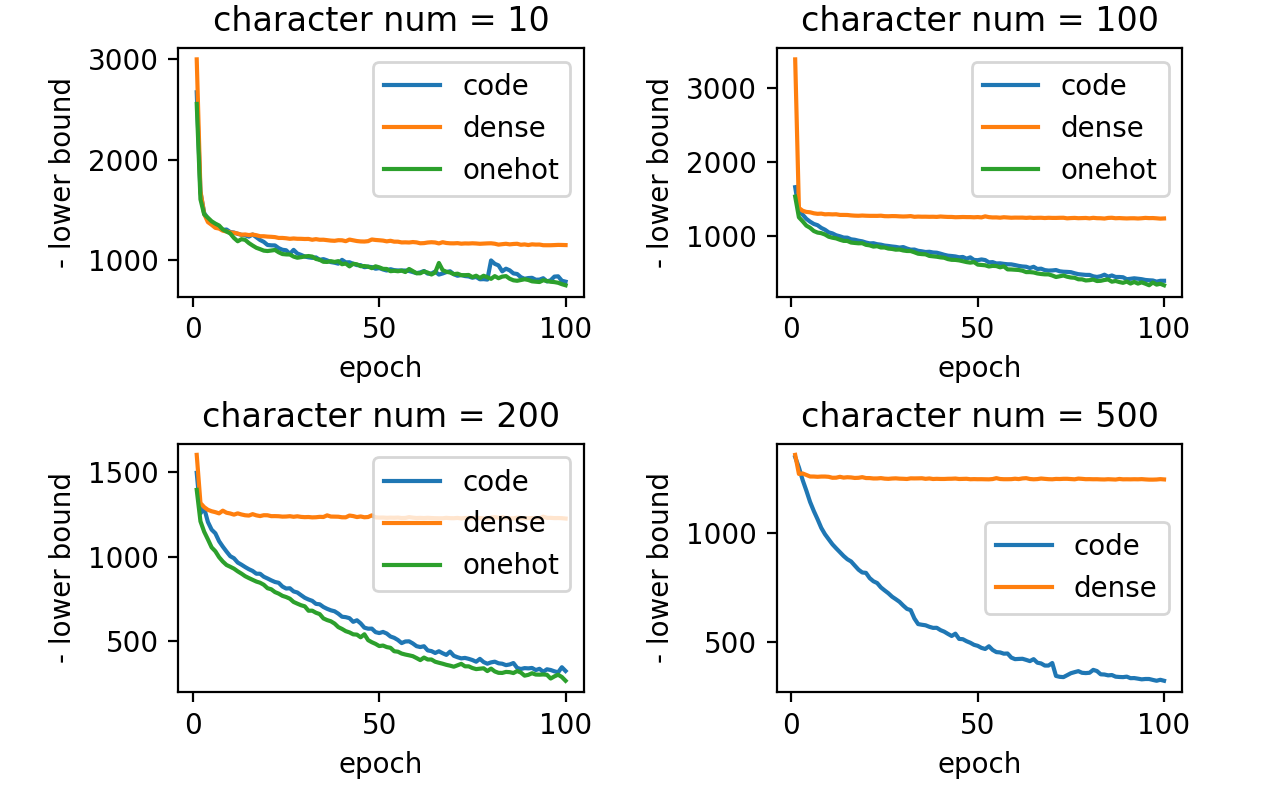}
    \caption{The training curves of three encoding methods using different numbers of unique characters, where the y-axis shows the negative lower bound and the x-axis shows the number of epochs during training. Notice that when we add the character numbers to 500, the model parameters using one-hot embedding will explode and we cannot show the corresponding curve.}
    \label{figure:plot}
\end{figure}

Our encoding method with domain knowledge instead of the one-hot embedding provides a solution for the large dictionary issue of Chinese characters. To show the effectiveness of knowledge guidance, we compare the converge speed of three encoding methods -- our content code, the one-hot embedding and the binary embedding. The results are shown in Fig.~\ref{figure:plot}.

Depending on the cost of expanding model capacity, the one-hot embedding may provide a comparable converge speed with our knowledgeable content code. However, this embedding method will make the model parameters exploded in a larger Chinese characters dictionary. The binary embedding consumes shorter length ($\lceil \log_2{N}\rceil$ bits) , but it hardly converges to a meaningful result. 

\section{Conclusions}
We have presented a novel Style-Aware VAE framework, which achieves the stylized Chinese character generation by reading only one or a few characters. Our model disentangles the style information and the content information with the intercross pair-wise optimization method and shows a powerful one-shot or few-shot generalization ability with new styles. Experiments demonstrate the impressive generation results on characters of both printing and handwritten styles. 



\section*{Acknowledgements}
The work is supported by the National NSF of China (Nos. 61571261, 61620106010, 61621136008, and 61332007), Beijing Natural Science Foundation (No. L172037), Tsinghua Tiangong Institute for Intelligent Computing and the NVIDIA NVAIL Program, and partially funded by Microsoft Research Asia and Tsinghua-Intel Joint Research Institute. Our project is built on ZhuSuan~\cite{shi2017zhusuan}, which is a deep probabilistic programming library based on Tensorflow.


\appendix
\section{Generation Results of Symbolic Alphabets}
\begin{figure}[!htb]
\centering
\subfigure[One-shot digit generation]{
\begin{minipage}[b]{\linewidth}
\centering
\includegraphics[width=0.55\linewidth]{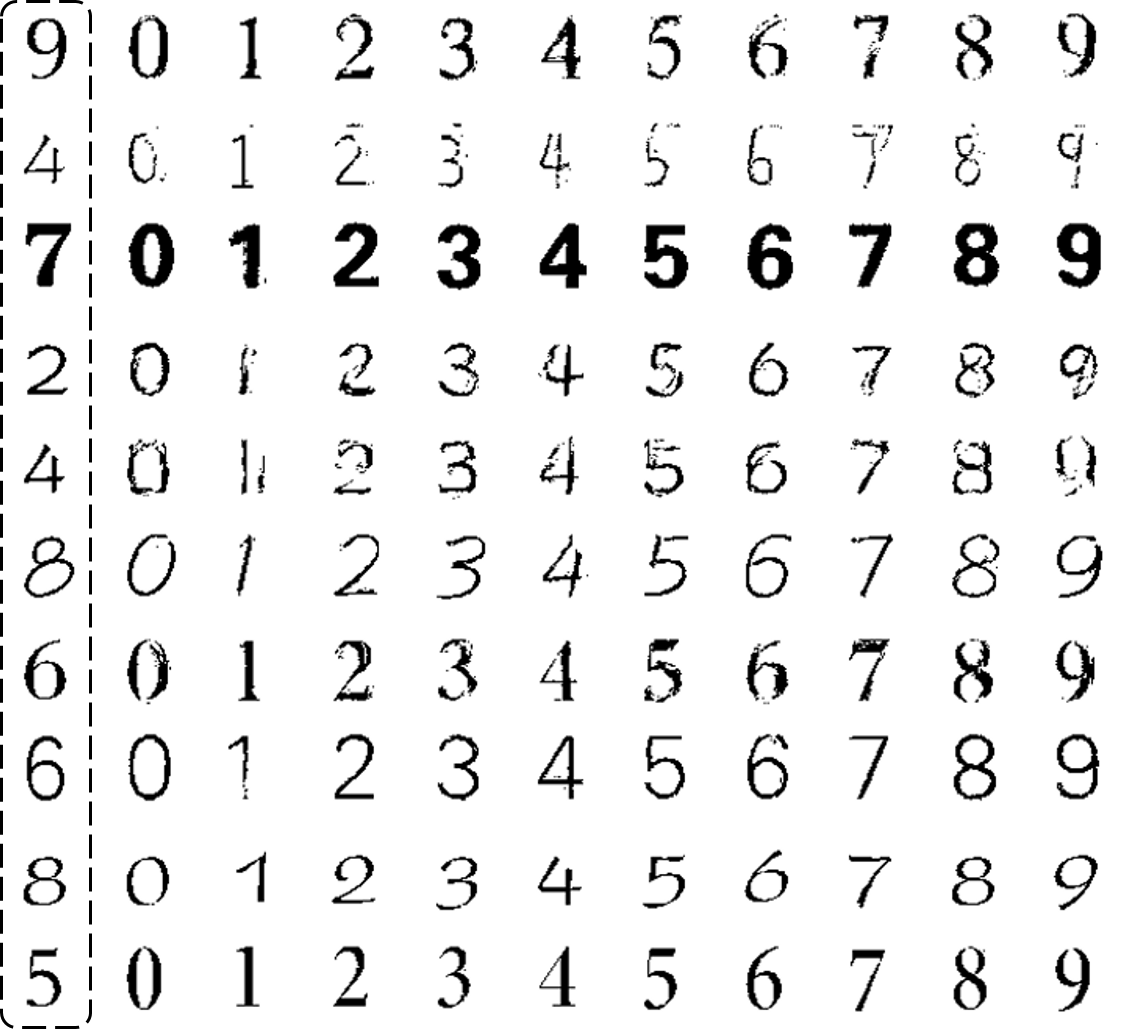}
\end{minipage}
}
\subfigure[One-shot English character generation]{
\begin{minipage}[b]{\linewidth}
\centering
\includegraphics[width=0.55\linewidth]{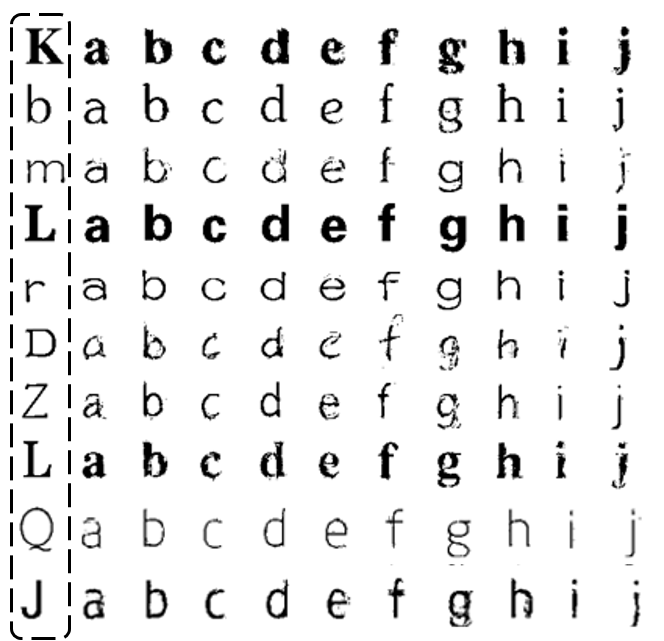}
\end{minipage}
}
\subfigure[One-shot English character generation]{
\begin{minipage}[b]{\linewidth}
\centering
\includegraphics[width=0.55\linewidth]{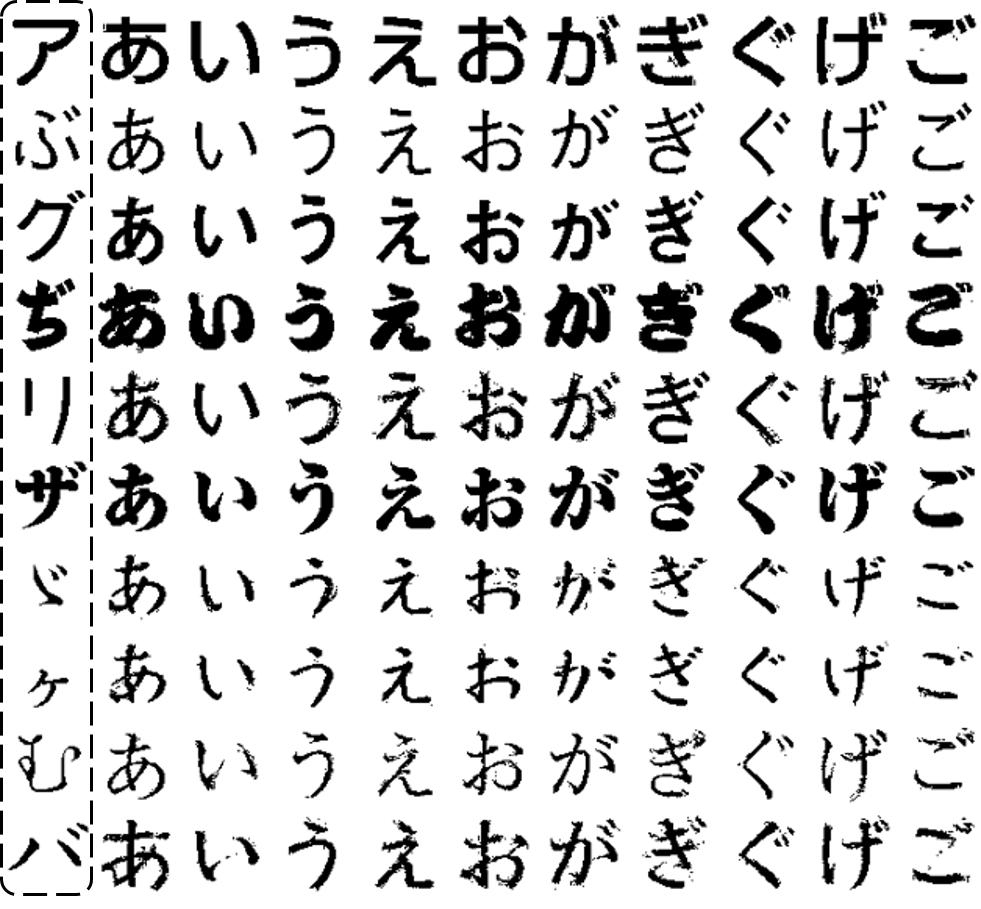}
\end{minipage}
}
\caption{The characters in the dotted frames are as the style providers and the others are generated characters in each row. As the size of their dictionary is not large, we use one-shot embedding to label each character. The style is consistent in each row.}
\label{fig:symbol}
\end{figure}

We also apply our SA-VAE framework for other simple symbolic characters including digits, English letters and Japanese kanas. The experimental results are shown in Fig.~\ref{fig:symbol}. 
Similar to the setting of Fig.~\ref{fig:fewshot}, the characters in the dotted frame are as the style providers and the others in the same row are the one-shot generation results with the specified style. Contrast to the Chinese character generation, we use one-hot embedding to label these symbolic characters because there is no complex structural information in them. The experimental results imply that our SA-VAE framework is not only restricted in Chinese character generation, but also a general technique for the disentanglement and one-shot generalization.

\bibliographystyle{named}
\bibliography{ijcai18}
\end{document}